\documentclass{article}
\usepackage{ijcai26}

\usepackage{times}
\usepackage{soul}
\usepackage{url}
\usepackage[hidelinks]{hyperref}
\usepackage[utf8]{inputenc}
\usepackage[small]{caption}
\usepackage{graphicx}
\usepackage{amsmath}
\usepackage{amsthm}
\usepackage{booktabs}
\usepackage{algorithm}
\usepackage{algorithmic}
\usepackage[switch]{lineno}
\usepackage{subcaption}
\usepackage{amsmath}
\usepackage{marvosym}
\usepackage{amssymb}

\usepackage{enumitem}
\usepackage{stfloats}

\usepackage{tikz}
\usepackage[edges]{forest}
\usetikzlibrary{trees,positioning,shapes,shadows,arrows.meta}

\usepackage{times}
\usepackage{latexsym}

\newcommand{\eat}[1]{}

\title{From Multimodal Perception to Strategic Reasoning: A Survey on AI-Generated Game Commentary}

\author{
{\fontsize{12}{14}\selectfont\bfseries
Qirui Zheng$^{1,*}$,
Xingbo Wang$^{1,*}$,
Keyuan Cheng$^{1,*}$,
Yunlong Lu$^{1}$,
Muhammad Asif Ali$^{2}$,
Lingfeng Li$^{1}$,
Yongyi Wang$^{1}$ {\normalfont and} Wenxin Li$^{1,\text{\Letter}}$}\\
\affiliations
$^1$Peking University\\
$^2$King Abdullah University of Science and Technology\\
\emails
zzzqr@stu.pku.edu.cn,
jacksimbol@stu.pku.edu.cn,
lwx@pku.edu.cn
}

\begin{document}

\maketitle

\begin{abstract}
The advent of artificial intelligence has propelled AI-Generated Game Commentary (AI-GGC) into a rapidly expanding research area, offering advantages such as scalable availability and personalized narration. However, existing studies remain fragmented, and a systematic survey that unifies prior efforts is still lacking.
To bridge this gap, our survey introduces a unified framework that systematically organizes the AI-GGC landscape. We present a novel taxonomy focused on three core commentator capabilities: Live Observation, Strategic Analysis, and Historical Recall, and further categorize commentary into three corresponding types: Descriptive Commentary, Analytical Commentary, and Background Commentary. Building on this structure, we provide an in-depth review of methods, datasets, and evaluation metrics, analyzing their strengths and limitations. Finally, we highlight key challenges and point out promising directions for future research in AI-GGC.
\end{abstract}


\section{Introduction}
\label{sec:intro}


Game commentary is the real-time narration and analysis of gameplay, which enhances the viewing experience by elucidating strategies and highlighting pivotal moments~\cite{review29}. Game commentary often appear in three major domains: board games (Chess, Go), sports (basketball, soccer), and esports (League of Legends (LoL), Dota 2). Its importance is evidenced by the massive viewership of major tournaments, such as the World Cup Final, which attract billions of concurrent viewers and rely heavily on commentary to sustain engagement and immersion.


The rapid advancement of AI has made \textit{AI-Generated Game Commentary} (AI-GGC) feasible. AI commentators present several advantages over human commentators. Primarily, they ensure unlimited availability, covering matches that cannot be covered by human commentators due to limited broadcast resources~\cite{review66}. Furthermore, they offer unparalleled personalization~\cite{review70}. Through tailored training, AI commentators can adapt to diverse viewer preferences, including the choice of language, voice, favored teams, and the level of analytical detail.

Although considerable research has been conducted, the field still lacks a systematic survey that consolidates these fragmented endeavors. This fragmentation stems primarily from two fundamental challenges inherent to the domain: 

\noindent{\bf (1) Strategic depth parsing.} In board games such as Chess and Go, while RL-based AI systems have achieved superhuman performance~\cite{alphago}, pure Large Language Models (LLMs) often struggle to provide deep, principled analysis of the reasoning behind each move. They can describe the moves made but frequently fail to articulate the underlying strategic reasoning and long-term planning behind them—reflecting their limited capacity for complex logical reasoning and long-horizon strategic inference.

\noindent{\bf (2) Multi-modal heterogeneity.} Different game genres essentially "speak different languages". Board games rely on simple, symbolic formats\eat{ like move notations}; sports games are captured as fast-paced, visually rich videos; while e-sports generate high-frequency, complex metadata streams. The entirely different techniques needed to process each format thus segregate research into isolated camps, preventing a unified methodology.

To address this gap, Section~\ref{sec:section2} introduces a systematic survey scheme for analyzing AI commentators, guided by two key questions: ``\textit{What makes a proficient commentator?}'' and ``\textit{What defines effective commentary?}'' We clarify the three core capabilities required of AI commentators and propose a taxonomy that classifies commentary into three types by content and purpose.
Building on this foundation, Section~\ref{sec:section3} reviews technical methodologies for these core capabilities. Section~\ref{sec:section4} surveys datasets by game genre and data practices, while Section~\ref{sec:section5} analyzes evaluation metrics. Figure~\ref{fig:taxonomy} presents our integrated taxonomy of methods, datasets, and metrics. Finally, Section~\ref{sec:section6} discusses open challenges and future directions, and Section~\ref{sec:section7} concludes the paper.
To summarize, this survey makes following key contributions:

\begin{enumerate}[leftmargin=0.5cm]
  \itemsep0em
    \item We propose a unified survey scheme that organizes AI-GGC along game genres, core commentator capabilities, and commentary types, providing a principled structure for analyzing existing methods and their trade-offs.
    \item Building on this scheme, we present the first comprehensive survey of AI-GGC spanning methods, datasets, and evaluation, and introduce a systematic taxonomy that synthesizes fragmented literature into a coherent framework.
    \item We critically examine the current AI-GGC landscape, highlighting challenges in capability coverage and maturity, system-Level coordination, and evaluation.
\end{enumerate}


\section{AI-GGC Survey Scheme}
\label{sec:section2}

\begin{figure*}
    \centering
    \includegraphics[width=0.95\linewidth]{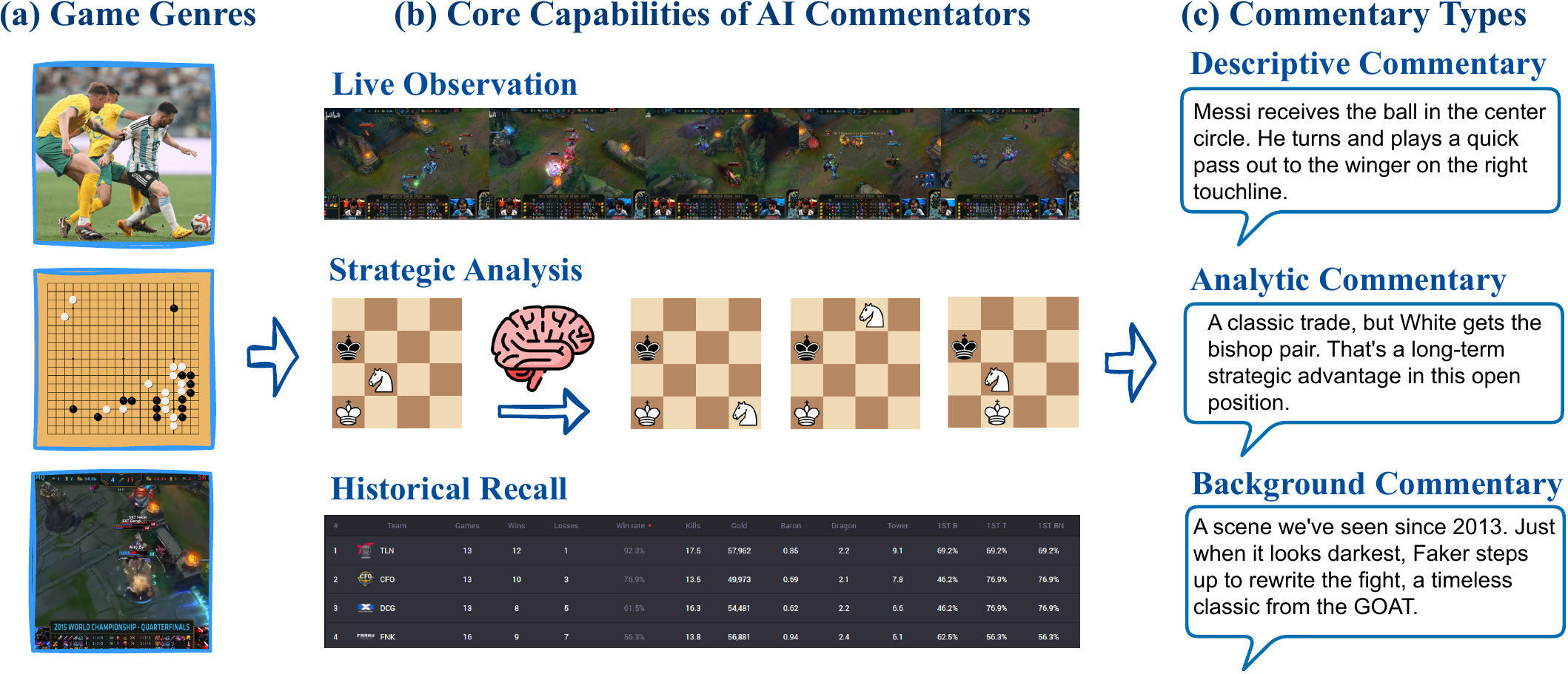}
    \caption{Overview of the proposed AI-GGC survey scheme, systematically summarizing the field along three key dimensions: (a) Game genres; (b) Foundational capabilities of AI commentators; (c) Commentary types.}
    \label{fig:scheme}
\end{figure*}

In this section, we propose a systematic survey scheme, illustrated in Figure~\ref{fig:scheme}, which organizes and summarizes the field of AI-GGC along three principal dimensions: game genre, foundational AI capabilities, and commentary types.

\subsection{Scope and Game Genres}
\label{subsec:2.1}

Our survey scheme is anchored in three representative and distinct game genres: \textbf{(1) Board Games.} Covers games with perfect information (Chess) or imperfect information (Mahjong). Commentary focuses on reasoning, move quality evaluation, and strategic analysis. \textbf{(2) Esports.} Includes fast-paced digital titles like Dota and LoL. Commentary analyzes real-time decision-making and both individual and team-based strategies. \textbf{(3) Sports.} Physical, dynamic games like soccer and basketball. Commentary emphasizes real-time movement, key events, and tactical coaching decisions.

In Section~\ref{sec:section4}, we categorize and analyze existing AI-GGC datasets according to these game genres.

\subsection{Core Capabilities of AI Commentators }
\label{subsec:2.2}

Drawing inspiration from the expertise of human commentators across various game genres, we abstract the core cognitive modules that underpin high-quality AI-GGC systems.

\textbf{(1) Live Observation (LO)} is responsible for the accurate perception and interpretation of real-time events and fine-grained details within the game environment. It serves as the primary data source for generating descriptive commentary, such as identifying the placement of pieces in chess, or describing shots and passes in soccer, kills and assists in LoL.

\textbf{(2) Strategic Analysis (SA)} encompasses the analytical abilities required to comprehend the current game state, interpret strategic intent, evaluate decisions, and anticipate future developments. It is particularly vital in board games, where understanding the intention and implications of each move is essential for insightful commentary.

\textbf{(3) Historical Recall (HR)} involves leveraging extensive game knowledge, including player histories, classic match moments, and community-specific terminology. It enables the commentator to provide meaningful context, such as recognizing signature moves or recalling iconic past plays, thereby enhancing the informativeness and engagement.

A systematic discussion of how existing methods implement these foundational capabilities is provided in Section~\ref{sec:section3}.

\subsection{Taxonomy of Commentary Types}
According to the three cognitive modules, we classify commentary types, which provides a new evaluation dimension.

\textbf{(1) Descriptive Commentary.} This is the most fundamental type, directly derived from the LO capability. It objectively addresses the question, \textit{“What is happening?”} by narrating immediate, observable events in games.

\textbf{(2) Analytical Commentary.} Derived from SA, this commentary transcends basic description by evaluating the quality of decisions or actions \textit{(“How good?”)}, interpreting player intent \textit{(“Why?”)}, and predicting strategic outcomes within a broader tactical context \textit{(“What’s next?”)}.

\textbf{(3) Background Commentary.} Enabled by the HR capability, this commentary type enriches the live narrative by integrating broader contextual information, such as detailed team and player backgrounds, game history, and cultural references. It addresses the question \textit{“What is the significance of this?”} by referencing past performances, historical match situations, team history, and community knowledge.

\eat{This taxonomy provides a systematic foundation for the development and evaluation of AI-GGC. By categorizing commentary into distinct functional types, the survey scheme facilitates a more rigorous examination of what constitutes high-quality commentary, recognizing that each category necessitates specific qualitative and quantitative assessment criteria.} 


\colorlet{hidden-draw}{black}
\tikzstyle{my-box}= [
    rectangle,
    draw=hidden-draw,
    rounded corners,
    text opacity=1,
    minimum height=1.5em,
    minimum width=5em,
    inner sep=2pt,
    align=center,
    fill opacity=.5,
]
\tikzstyle{leaf}=[my-box, minimum height=1.5em,
    fill=blue!15, text=black, align=left,font=\large,
    inner xsep=2pt,
    inner ysep=4pt,
]

\begin{figure*}[t]
    \centering
    \resizebox{\textwidth}{!}{
        \begin{forest}
            forked edges,
            for tree={
                grow=east,
                reversed=true,
                anchor=base west,
                parent anchor=east,
                child anchor=west,
                base=left,
                font=\Large,
                rectangle,
                draw=hidden-draw,
                rounded corners,
                align=left,
                minimum width=2em,
                edge+={darkgray, line width=1pt},
                s sep=2pt,
                inner xsep=2pt,
                inner ysep=3pt,
                ver/.style={rotate=90, child anchor=north, parent anchor=south, anchor=center},
            },
            where level=1{text width=6em,font=\Large,, align=left}{},
            where level=2{text width=8.9em,font=\large,}{},
            where level=3{text width=6.4em,font=\large,}{},
            where level=4{text width=6.4em,font=\large,}{},
        [ \;\;AIGGC\;\; , ver
                [\;Methods\\\;\;\;\;(\S \ref{sec:section3})
                    [\;Live Observation (\S \ref{subsec:3.1}), text width=11.6em
                        [\;Structured Text, text width=11.2em
                            [
                                \;Forsyth–Edwards Notation; Extended Position Description; Smart Game Format.
                                ,leaf ,text width=45.5em]]
                        [\;Event Spotting, text width=11.2em
                            [
                                ~\cite{review36,review49,review56,review94}.
                                ,leaf ,text width=45.5em]]
                        [\;Direct Encoding, text width=11.2em
                            [
                                ~\cite{review29,review26,review17,review18}.
                                ,leaf ,text width=45.5em]]
                    ]
                    [\;Strategic Analysis (\S \ref{subsec:3.2}),text width=11.6em
                        [\;Prompt-based CoT, text width=11.2em
                            [
                                ~\cite{review58,review3}.
                                ,leaf ,text width=45.5em]]
                        [\;Externally Augmented \\\;Reasoning, text width=11.2em
                            [\;Human Expertise, text width=8.1em
                            [
                                ~\cite{review56,review58,review3}; \\
                                ,leaf ,text width=35.8em]]
                            [\;Game AI, text width=8.1em
                            [
                                ~\cite{review26,review24,review35,review26}
                                ,leaf ,text width=35.8em]]
                                ]
                    ]
                    [\;Historical Recall (\S \ref{subsec:3.3}),text width=11.6em
                        [
                            ~\cite{baseballstory,review70}.
                        ,leaf ,text width=58.1em]
                    ]
                    [\;Content Organization \\\;(\S \ref{subsec:3.4}),text width=11.6em
                        [
                            ~\cite{review1,review63,review49}.
                        ,leaf ,text width=58.1em]
                    ]
                    [\;Commentary Generation\\\;and Rendering (\S \ref{subsec:3.5}),text width=11.6em
                        [\;Generation Methods, text width=11.2em
                            [\;Template-based, text width=8.1em
                            [
                                ~\cite{review33,review44,review21,review39}.
                                ,leaf ,text width=35.8em]]
                            [\;DL-based, text width=8.1em
                            [
                                ~\cite{review17,review73,review31,review63}.
                                ,leaf ,text width=35.8em]]
                            [\;LLM-based, text width=8.1em
                            [
                                ~\cite{review94,review2,review70}.
                                ,leaf ,text width=35.8em]]
                                ]
                        [\;Output Rendering, text width=11.2em
                            [
                                ~\cite{review39,review66}.
                                ,leaf ,text width=45.5em]]
                    ]
                ]
                [\;Datasets \\\;\;\;\;(\S \ref{sec:section4})
                    [\;Board Game \\\;Commentary Dataset\\\;
                    (\S \ref{subsec:4.2}),text width=11.6em
                        [\;Representative Datasets, text width=11.2em
                            [
                                \;Chess~\cite{review29}; Shogi~\cite{review69}; Go~\cite{review64}.
                                ,leaf ,text width=45.5em]]
                        [\;Specially Annotated, text width=11.2em
                            [
                                \;Shogi Commentary Corpus~\cite{review88}.
                                ,leaf ,text width=45.5em]]
                    ]
                    [\;Sports Commentary\\\;Dataset (\S \ref{subsec:4.3}),text width=11.6em
                        [\;Representative Datasets, text width=11.2em
                            [
                                \;SoccerNet series~\cite{review92,review90,soccernetv3}; \\
                                \;GOAL~\cite{review95}; SVN~\cite{review31}; FSN~\cite{fsn}; \\
                                \;SoccerTrack-Commentary Dataset~\cite{vijayakumar2025player}; OptaSports~\cite{review52}.
                                ,leaf ,text width=45.5em]]
                        [\;Automatic Alignment, text width=11.2em
                            [
                                \;MatchTime~\cite{review2}.
                                ,leaf ,text width=45.5em]]
                        [\;Specially Alignment, text width=11.2em
                            [
                                \;Soccernet-Echos~\cite{review4}; SoccerNet-Caption~\cite{review1}.
                                ,leaf ,text width=45.5em]]
                        [\;Story Resources, text width=11.2em
                            [
                                \;Cricket Story Library~\cite{story}; Baseball Story \;Dataset~\cite{review34}.
                                ,leaf ,text width=45.5em]]
                    ]
                    [\;Esports Commentary\\\;Dataset (\S \ref{subsec:4.4}),text width=11.6em
                        [\;Representative Datasets, text width=11.2em
                            [
                                \;Game-MUG~\cite{review5}; Dota Commentary Dataset~\cite{review77}.
                                ,leaf ,text width=45.5em]]
                        [\;Multimodal Datasets, text width=11.2em
                            [
                                \;LoL-V2T~\cite{review76}; CS-LoL~\cite{cslol}.
                                ,leaf ,text width=45.5em]]
                    ]
                ]
                [\;Evaluation\\\;Metrics \\\;\;\;\;(\S \ref{sec:section5})
                    [\;Human-Centric Metrics, text width=11.6em
                        [\;Likert-scale~\cite{review73,review94,review52,review56}.  ,leaf ,text width=58.1em]
                    ]
                    [\;Untrained Automatic\\\;Metrics, text width=11.6em
                        [\;N-gram Overlap Metrics, text width=11.2em
                            [{\;BLEU~\cite{bleu}; BLEU-2, cBlEU, sCLEU~\cite{modifiedBleu}; \\
                            \;ROUGE-1 / 2 / 3 / 4 / L~\cite{rouge}; METEOR~\cite{meteor}.}, leaf, text width=45.5em]
                        ]
                    ]
                    [\;ML-based Metrics, text width=11.6em
                        [\;BERTScore~\cite{review5}; SnowNLP~\cite{review3}.
                        ,leaf ,text width=58.1em
                        ]
                    ]
                    [\;LLM-based Metrics, text width=11.6em
                        [\;GCC-Eval~\cite{geval,review58}; Qwen2.5-VL-72B-Instruct~\cite{you2025timesoccer}
                        ,leaf ,text width=58.1em
                        ]
                    ]
                ]
        ]
        \end{forest}
    }
    \caption{Taxonomy of Methods, Datasets and Metrics in AI-GGC}
    \label{fig:taxonomy}
\end{figure*}

\section{Methods}
\label{sec:section3}

This section reviews technical pathways towards core capabilities of AI commentators introduced in Section~\ref{sec:section2}. We also expand our discussions to architectural designs that wire key function modules to produce end-to-end game commentaries.

\subsection{Live Observation}
\label{subsec:3.1}

LO requires AI-GGC systems to process complex, multimodal game data under time constraints. Board games typically rely on direct visual observation of game board or structured textual representation that encode game state into compact strings (e.g. \textit{Forsyth-Edwards Notation} for chess), while sports and e-sports rely mostly on videos.
However, explaining “what is going on” in fast-paced games involves more than multimodal perception. A good commentary should be able to identify focal points in chaotic gameplays and summarize them using professional terminology, balancing detail and conciseness under time and comprehension constraints 


Early technical trials attempting to address this challenge primarily adopt \textbf{Event Spotting (ES)}, which reduces the complexity of raw game video processing by converting visual inputs into textual labels.
Specifically, predefined event labels
are assigned to corresponding timestamps in game videos and subsequently used as inputs to the commentary generation model, instead of raw video streams.
Early works~\cite{review92,review90} rely on human-annotated domain-specific datasets to train models for ES which may suffer from generalization issues. Detection-based ES reduces annotation labor by using object-detection models to identify actions and entities in videos. Jung~\shortcite{jung2025integrated} detect players and the ball using YOLOv8 and perform ES by combining spatial-temporal trajectories, action recognition, and rule-based reasoning to identify key basketball events. Similar pipelines have also been explored using various object detection backbones~\cite{review49}. 

While effective in reducing perceptual complexity, this representation inevitably introduces information loss by discarding fine-grained details beyond the detected events.

\textbf{Direct Encoding} approaches map multimodal inputs (e.g. video streams and metadata) into dense representations in order to preserve information completeness. 
The advancing multimodal large language models has made it possible. Initial efforts focused on encoding structured representations for board games, directly encoding chessboard states into vectors~\cite{review29,review26}, until recent advance in models' multimodal understanding capabilities enable the extension of this approach to visual inputs from sports. Jhamtani~\shortcite{review18} combined video transformers for visual input and BERT for textual descriptions, fusing these features with a transformer encoder guided by a soft-prompt mechanism for better multimodal perception.

Despite recent progress, several direct encoding approaches still rely on relatively early visual encoding backbones~\cite{review8}, which may limit their capabilities for modeling complex visual information. 
More fundamentally, recent studies show that even state-of-the-art LLMs struggle with visual perception in low-dynamic board games~\cite{wang2025large},raising concerns about their scalability to fast-paced and highly dynamic sports or esports.

\begin{figure}[t]
    \centering
    \begin{subfigure}[t]{0.52\linewidth}
        \centering
        \includegraphics[width=\linewidth]{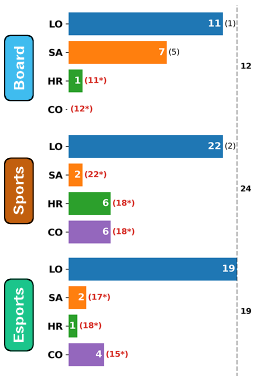}
        \caption{Methods count of AI-GGC systems across core commentator capabilities, grouped by game genres.} 
        \label{fig:method1}
    \end{subfigure}
    \hfill 
    \begin{subfigure}[t]{0.44\linewidth}
        \centering
        \includegraphics[width=\linewidth]{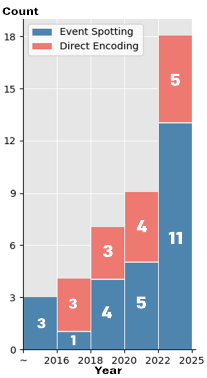}
        \caption{Temporal distribution of LO method counts in AI-GGC.}
        \label{fig:method2}
    \end{subfigure}
    \caption{Methodological patterns in AI-GGC research across capabilities and time. All statistics are reported by paper count.} 
    \label{fig:method}
\end{figure}

Methodologically, LO paradigms in AI-GGC have evolved in response to changing modeling constraints and opportunities, as reflected in the temporal distribution of methods in Figure~\ref{fig:method2}. 
Early work almost exclusively adopted ES to simplify video modeling under limited capacity, inevitably introducing information loss. 
With advances in DL, direct encoding emerged as a parallel paradigm between 2016 and 2022, aiming to preserve richer visual information. 
Since 2023, the rise of LLMs has renewed interest in event-based representations, as discrete events can be readily transformed into textual or structured prompts without resolving visual perception, making ES attractive from an engineering perspective. 
Importantly, these trends do not imply that direct encoding is a dead end. Rather, its practicality remains contingent on advances in multimodal understanding, suggesting that it may regain relevance as model capabilities continue to evolve.


\subsection{Strategic Analysis}
\label{subsec:3.2}

A key challenge for AI-GGC is enabling deep SA, a capability that places substantial demands on the logical reasoning ability. Beyond LO, a good commentator is expected to address \textit{“How good”}, explain \textit{“Why”}, and anticipate \textit{“What’s next”} within the same logic. Depending on whether SA relies solely on the model itself, we categorize existing AI-GGC methods for SA into \textit{Prompt-based Chain-of-Thought(CoT)} and \textit{Externally Augmented Reasoning}~\cite{Li2025PerceptionRT}.

\textbf{Prompt-based CoT} refers to eliciting step-by-step SA purely through instructions in the prompt, guiding the model to carry out deeper reasoning without introducing external modules, search procedures, or additional training. In AI-GGC, prompting is not only used to steer how the model analyzes the current situation, but also to inject details of game rules and gameplay common sense that LLMs may be unfamiliar with (e.g., in Guandan, three-with-a-pair is allowed whereas three-with-a-single is not).
Prior work~\cite{review58,review3} has shown that such prompt-based guidance can better elicit SA ability of LLMs and improve the quality of generated commentary.

Despite its simplicity and effectiveness, this method mainly elicits SA already latent in the model, without extending its intrinsic reasoning capacity. As a result, its ability to support deeper strategic inference and forward-looking anticipation remains bounded by the model itself.

\textbf{Externally Augmented Reasoning} refers to enhancing SA by introducing auxiliary components beyond the language model itself, such as advanced algorithms or expert knowledge, to compensate for limited reasoning capacity of models. Such augmentation typically takes the form of external guidance, which can be broadly categorized into guidance derived from \textit{Human Expertise} and guidance provided by \textit{Game AIs}.

Incorporating strategic principles from \textbf{Human Expertise} allows LLMs to perform more insightful analysis, analogous to expert commentators explaining key moves and tactical intentions. The main challenge lies in embedding such principles effectively into the model’s reasoning process. Kim~\shortcite{review56} addressed this by manually mapping baseball events to strategic concepts and provided both as model inputs, enabling interpretation within an explicit strategic context. In chess, Kim~\shortcite{review58} extracted key strategic concepts (e.g., king safety) using a game engine and compared their representations before and after each move to identify salient strategic changes for commentary generation. For imperfect-information games such as GuanDan, Ma~\shortcite{review3} proposed a two-level prompting scheme grounded in Theory of Mind (ToM)~\cite{frith2005theory}, where one prompt guided the model to analyze its own strategy while the other encouraged inference over opponents’ possible hands and intentions.

Augmenting strategy analysis with powerful \textbf{Game AIs} is another natural idea, mirroring common practice among human commentators, particularly in some board games where game AIs dominant. However, this approach introduces additional challenges. Despite their strength, RL-based game AIs are often black boxes, limiting interpretability and making it difficult to directly guide logical reasoning. Consequently, existing approaches focus on transforming unexplainable outputs into more explainable signals.
An early idea is to achieve this through controllable search or rule-based mechanisms. Kameko~\shortcite{review35} employed rule-based scoring functions and limited-depth search in Shogi to generate candidate game trees andF derive commentary from simulated outcomes.
Another idea is to translate engine outputs into interpretable linguistic signals. Lee~\shortcite{review24} quantified “move quality” by measuring win rate difference between the played move and the AI-recommended best move, and this difference was mapped to discrete natural-language categories. Zhang~\shortcite{review26} jointly trained a chess engine and a language model in a multi-task framework, allowing the system to simultaneously learn game understanding and commentary generation.

Although externally augmented reasoning substantially improves the logical coherence and correctness of analytical commentary, it also faces inherent limitations. First, both sources of external guidance exhibit inherent weaknesses. Human expertise provides only partial coverage of the strategic space, generalizes poorly to complex or unseen situations, and may become outdated over time. Game AI–based augmentation, while powerful, is constrained by the black-box nature of modern engines; simple mappings from engine outputs to linguistic labels are often insufficient for rich strategic explanations.
More critically, in highly complex sports and esports domains, dominant game AIs are still largely unavailable, severely limiting the applicability of AI-driven augmentation. Second, external augmentation introduces capabilities absent from the model itself, but these remain externally provided rather than internalized as intrinsic reasoning ability.

Beyond method-specific limitations, existing AI-GGC approaches share a fundamental weakness in SA: they mainly focus on short-horizon quality assessment (“How good?”), without constructing coherent multi-step strategic narratives. As a result, reasoning about \textit{“Why?”} is often incomplete, leading to unreliable prediction of \textit{“what's next”} based on such ligic. And as also discussed above, current approaches rarely internalize strategic reasoning into the model itself, instead mainly relying on externally provided signals.

\subsection{Historical Recall}
\label{subsec:3.3}

Historical recall enriches background commentary by integrating external knowledge. Previous works have approached this primarily as an information retrieval task. Lee~\shortcite{baseballstory} ranked stories from a database based on their similarity to the current game state using AdaRank~\cite{adaRank}, then filtered candidates with a regression model before inserting into commentary. Andrews~\shortcite{review70} encoded player information with SBERT to retrieve relevant player or team data.

\subsection{Content Organization}
\label{subsec:3.4}

While techniques in former sections have enabled rich contents and details based on raw game states and data, real game commentary is usually restricted by limited time and audiences' capabilities to absorb the game. Thus AI-GGC faces an inherent challenge: transforming unstructured segments into logically structured, concise narratives through strategic selection and prioritization to maintain viewer comprehension without cognitive overload. Mkhallati~\shortcite{review1} proposed a spotting head that predicts when commentary should be generated for each time window, using non-maximum suppression to eliminate redundancy. Mei~\shortcite{review63} introduced a coarse-to-fine aligner that filters out irrelevant records and reweights alignment scores to focus on salient events. Likewise, Ranella~\shortcite{review49} used an event-priority and delayed-confirmation mechanism to filter key events in real time.

\subsection{Commentary Generation and Rendering}
\label{subsec:3.5}

Commentary generation, the final stage of AI-GGC, transforms processed information into natural language. These methods fall into three main categories: template-based, DL-based, and LLM-based. Template-based approaches are efficient and reliable~\cite{review33,review44,review21}, but often yield repetitive outputs. DL-based methods, typically using LSTM decoders~\cite{review73,review31,review63}, provide more diverse and context-aware commentary. LLMs further improve fluency and strategic reasoning, with prompt-based techniques explored in recent work~\cite{review94,review70,review58}, though they can be slower, more costly, and prone to hallucinations.

Multimodal rendering is an emerging area, with most non-textual commentaries generated by converting text to speech via TTS APIs. For example, Xu~\shortcite{review39} generated both text and audio parameters for expressive speech, while Siu~\shortcite{review66} combined TTS and StyleGAN to create a virtual commentator with realistic facial expressions and lip-syncing.

\subsection{A Systematic Gap in Capability Coverage}
\label{subsec:3.6}
Figure~\ref{fig:method1} reveals a pronounced structural gap in current AI-GGC research: most systems focus on LO and fail to cover the full set of core commentator capabilities, with comparatively limited attention to SA, HR, and CO. This imbalance manifests differently across game genres due to their distinct characteristics. For board games, longer turn intervals and high demands for logical reasoning lead to greater research attention on SA, while reducing the relative need for CO. Powerful game AIs provide strong support for this ability, reinforcing this focus. In contrast, the limited attention to HR is difficult to justify. For sports and esports, the dominance of LO is largely driven by the intrinsic difficulty of multimodal perception in fast-paced and visually complex scenes, leaving other capabilities substantially under-explored despite their importance for effective commentary.


\section{Datasets and Benchmarks}
\label{sec:section4}
This section reviews dataset according to game gernes, along with relevant data processing techniques.






\subsection{Board Game Commentary Datasets}
\label{subsec:4.2}

Commentary datasets for board games have been developed for chess, Go, Shogi (Japanese chess), GuanDan (Chinese card game). The turn-based nature of board games ensures natural alignment between game states and commentary, and the slower pace offers sufficient time for deep reasoning.

Most datasets consist of aligned pairs of game states and commentaries. Jhamtani ~\shortcite{review29} introduced a large chess dataset with 298K move-by-move annotations, establishing a benchmark for AI-GGC in chess. Additional datasets exist for Shogi~\cite{review35} and Go~\cite{review64}.

Some datasets are annotated beyond commentary text for better commenting. A Shogi commentary corpus~\cite{review88} annotates modality expressions and event factuality, enabling fine-grained modeling of uncertainty and future-oriented reasoning in strategic commentary.

\subsection{Sports Commentary Datasets}
\label{subsec:sectionsport}
\label{subsec:4.3}
As mentioned before, in sport games, ES is widely used to simplify percetion of complex multimodal input.
ES-based datasets cover soccer~\cite{review95}, basketball~\cite{review31}, baseball~\cite{review56}\eat{, cricket~\cite{review74}}, and tennis~\cite{review73}. SoccerNet~\cite{review92} is a prominent example, providing soccer videos with event labels and over 500K commentaries. It expanded to SoccerNet-v2~\cite{review90} with 17 event categories and SoccerNet-v3~\cite{soccernetv3} with spatial annotations. To compensate for the information loss caused by ES, some datasets also incorporate additional modalities such as player tracking~\cite{vijayakumar2025player} and play data~\cite{review52}, enabling richer perception and reasoning. 

Manual alignment of game states and commentary remains common but is labor-intensive. To automate alignment, Rao~\shortcite{review2} developed a pipeline combining LLaMA-3 and contrastive learning, trained on a small set of aligned data.

Alternative alignment strategies have been introduced for specific tasks, including timestamp-based sparse alignment to support long-video commenting~\cite{review1} and benchmarks where commentaries outnumber events, reflecting real-world settings~\cite{review41}.

Beyond commentary datasets, some auxiliary resources can be used to enrich background commentary by providing anecdotes, such as story libraries for cricket~\cite{story} and baseball~\cite{baseballstory,review34}.

\subsection{Esports Commentary Datasets}
\label{subsec:4.4}
Esports datasets, like those for sports, often use ES to manage complex video inputs. A unique advantage in esports is the availability of official APIs, which provide precise access to rich multimodal signals such as player statistics, game state metadata, and audio streams~\cite{review11}, placing higher demands on multimodal perception.

Esports commentary datasets cover games such as LoL~\cite{review65}, Dota 2~\cite{review77}, and RoboCup~\cite{review62}, and often include rich multimodal data including audio features~\cite{review5}, sensor data~\cite{review32}, or multiple viewpoints~\cite{review67}. Ringer~\shortcite{review11} introduced affect labels capturing both game context and emotional dimensions, enabling affect-aware commentary generation.

\section{Evaluation Metrics}
\label{sec:section5}

Evaluating AI-GGC systems is inherently challenging for three reasons: 
(1) AI-GGC is a multi-faceted task involving various core commentator capabilities, with no clear way to aggregate them into a single score. (2) It is an open-ended NLG task with no gold reference, as different commentators may produce diverse yet equally valid commentaries for the same game state. (3) Analytical commentary often requires long chains of strategic reasoning that are difficult to formalize, making correctness hard to assess with existing metrics. 
Current AI-GGC work relies on evaluation methods that only partially address these challenges and have substantial limitations, which can be broadly grouped into four categories:

\textbf{Human-Centric Metrics.} Human-centric metrics remain the gold standard for assessing commentary quality, typically using questionnaires like Likert-scale. This metric of course can ignore the three challenges due to its directly reflecting human preference, but they are expensive, require domain expertise, and highly subject to individual variability.


\textbf{Untrained Automatic Metrics.}
Untrained automatic metrics, especially n-gram overlap measures, remain the most widely used tools for AI-GGC evaluation.
As shown in Figure~\ref{fig:method1}, there are nearly half of AI-GGC works using it as the only evaluation metric.
Metrics such as BLEU~\cite{bleu}, ROUGE~\cite{rouge}, and METEOR~\cite{meteor} quantify lexical similarity to reference texts and are popular due to their low cost and reproducibility.
However, n-gram overlap metrics are inherently misaligned with the task and correlate weakly with human judgments because AI-GGC lacks definitive ground-truth references~\cite{bleubad}, making them ineffective for addressing any of the three challenges discussed above.

\begin{figure}
   \centering
    \includegraphics[width=0.95\linewidth]{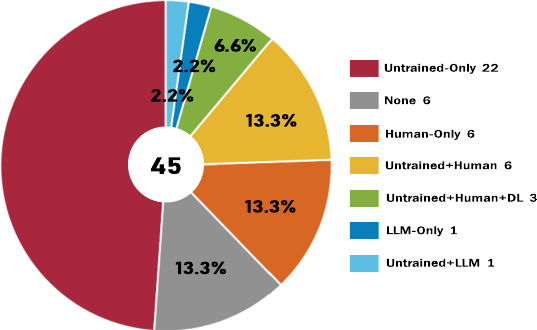}
    \caption{Distribution of evaluation strategies used in AI-GGC.}
    \label{fig:metrics}
\end{figure}

\textbf{ML-based Metrics.}
ML-based metrics instead use learned models to approximate semantic similarity or other properties of commentary.
Zhang~\shortcite{review5} use BERTScore to evaluate esports commentary. However, it is still a reference-based measure of similarity to human texts and thus shares the same fundamental limitation as n-gram metrics in AI-GGC.
Worth mentioning, Ma~\shortcite{review3} apply SnowNLP, a Naïve Bayes classifier, to estimate sentiment polarity of commentary.
Although it is a very coarse signal and should not be treated as a primary metric, it has limited value as an auxiliary indicator of whether the overall emotional tone is plausible.

\textbf{LLM-based Metrics}
move beyond simple similarity measures by providing reference-free, multi-dimensional assessments that better align with human preferences through prompting
or post-training.
For instance, Kim~\shortcite{review58} introduced GCC-Eval, which integrates expert chess knowledge with G-Eval~\cite{geval} to evaluate chess commentary along four dimensions: relevance, completeness, clarity, and fluency. 
However, it applies the same dimension weights to all commentary types and avoids assigning reasoning correctness to the LLM, relying instead on human annotations.
Although still an emerging approach, LLM-as-a-judge currently appears to be the most promising direction for evaluating AI-GGC.
However, existing work also shows that such judges remain unstable when verifying long chains of reasoning and exhibit systematic biases, such as a tendency to prefer longer~\cite{ye2024justice} and position bias~\cite{shi2024judging}, which limits their current reliability for AI-GGC.


Overall, existing AI-GGC evaluation remains only partially aligned with the task: untrained and ML-based metrics struggle with open-endedness and multidimensionality, while LLM-based judges are still unreliable for assessing long-chain reasoning. To address these limitations, we argue that the taxonomy of descriptive, analytical, and background commentary offers a natural basis for type-specific evaluation, as discussed in Section~\ref{sec:section6}.

\section{Challenges and Future Directions}
\label{sec:section6}
%

AI-GGC is rapidly emerging as a prominent application area for LLMs, where technical innovation meets growing real-world demand. While recent research and practical deployments highlight the field’s promise, they also expose fundamental challenges and point to key directions for future work.

\textbf{Incomplete Coverage of Core Commentator Capabilities.}
A common limitation of existing AI-GGC systems is their incomplete coverage of core AI commentator capabilities, as shown in Figure~\ref{fig:method1} and discussed in Section~\ref{subsec:3.5}.
\textit{Future AI-GGC research should therefore place balanced emphasis on all core commentator capabilities, treating them as equally important components in building more informative, interpretable, and complete commentary systems.}

\textbf{Shallow Realization of LO.} 
LO requires AI commentators to accurately capture and summarize ongoing games to guide audience attention and comprehension. Although a considerable amount of work has already been devoted to processing raw game data and extracting fundamental game events, these descriptions are still far from clear, specific, and easily understandable game descriptions. While pattern-recognition-based processing relies on models trained on annotated data with predefined event labels~\cite{jung2025integrated}, this class of methods often lead to significant information loss and preclude the possibility of subsequent fine-grained scene and strategy analysis. \textit{We advocate that the introduction of common sense and reasoning could pave the way towards ideal LO commentaries. On capturing atomic actions of game entities, using generic object detection techniques, the model can perform reasoning by:
(1) constructing relational graphs among multiple entities,
(2) conducting cross-temporal alignment with observation history, and
(3) referring to game-specific knowledge and commonsense,
thereby yielding fine-grained, comprehensive cognition of the current game state.}

\textbf{Shallow Realization of SA.}
Current AI-GGC systems exhibit several critical limitations in SA.
First, most systems lack long-horizon match memory, relying primarily on current actions and states for analysis, which makes it difficult to explain “Why?” or reliably anticipate “What’s next?”, both of which depend on broader historical context.
\textit{Maintaining memory of past actions and intermediate analytical states could help models form more coherent strategic interpretations.}
Second, the modeling space of external augmentation remains underexplored.
\textit{Explainable AI methods provides a direct way to understand black-box game AIs, particularly interpretable game AI models~\cite{li2025mxplainer}.}
\textit{Meanwhile, deeper ToM modeling can help LLMs reason about players’ intentions and beliefs~\cite{von2017minds}.}
Finally, SA is rarely internalized within LLMs, with most systems relying on external analytical cues, constraining reasoning depth and consistency.
\textit{Aligning LLMs with strategic reasoning behaviors of game AIs at the parameter level is a promising direction for more robust analytical commentary.}

\textbf{Lack of Cross-Capability Coordination.}
A common limitation of existing AI-GGC systems is the lack of coordination across core commentator capabilities.
LO, SA, and HR are often treated as independent modules rather than as an integrated whole, with limited communicating or joint decision-making.
Consequently, capabilities are optimized isolately, resulting in fragmented understanding and inconsistent commentary logic.
While You~\shortcite{you2025timesoccer} adopts an end-to-end formulation, it primarily couples LO with commentary generation, leaving other core capabilities uncoordinated.
\textit{Cross-capability coordination can be promoted from both informational and functional perspectives.
At the informational level, intermediate information produced by different capabilities can be partially shared.
At functional level, SA and HR can impose fine-grained information demands on LO, specifying which details should be emphasized.
Such demand-driven coordination remains unexplored in existing AI-GGC systems.
}

\textbf{Lack of a Standardized Benchmark and Reliable Evaluation.}
Despite the growing number of AI-GGC datasets spanning multiple modalities, game genres, and languages, the field still lacks a unified and standardized benchmark, making fair comparison across AI-GGC works difficult.
Moreover, as discussed in Section~\ref{sec:section5}, this problem is further exacerbated by the lack of reliable and convincing evaluation metrics.
\textit{A pressing direction is to develop a large-scale, multimodal AI-GGC benchmark with standardized data formats and fair evaluation.
For evaluation, fine-tuning preference models from human labels is feasible but costly.
In contrast, the taxonomy of commentary types proposed in our work enables more targeted LLM-as-a-Judge designs for different commentary types, alleviating challenges of multi-dimensionality and open-endedness.
The relative importance and weights of evaluation dimensions can be examined through expert assessment or audience feedback~\cite{valdivia2025evaluating}.
For analytical correctness, a promising direction is to decompose long reasoning chains into shorter, verifiable sub-chains and validate them progressively, reducing reliance on LLMs’ long-horizon reasoning evaluation.
}

\section{Conclusion}
\label{sec:section7}
In this survey, we provide a systematic and comprehensive overview of AI-GGC.
We propose a unified survey scheme that organizes existing work by game genres, core commentator capabilities, and commentary types, and use it to analyze methods, datasets, and evaluation metric.
Moreover, we highlight key research gaps in current AI-GGC works and identify several promising directions for future work.
We hope this survey offers a clear picture of the field and helps guide future research toward more mature AI commentators.



\section*{Contribution Statement}
Qirui Zheng, Xingbo Wang, and Keyuan Cheng contributed equally to this work. 
Wenxin Li is the corresponding author.






\bibliographystyle{named}
\bibliography{ijcai26}

\end{document}